\documentclass[10pt,twocolumn,letterpaper]{article}

\usepackage[pagenumbers]{iccv}      

\definecolor{iccvblue}{rgb}{0.21,0.49,0.74}
\usepackage[pagebackref,breaklinks,colorlinks,allcolors=iccvblue]{hyperref}
\usepackage{multirow}
\usepackage{xcolor}

\definecolor{mygreen}{RGB}{0,128,128} 
\definecolor{myred}{RGB}{178,34,34}   
\def\paperID{11471} 
\def\confName{ICCV}
\def\confYear{2025}

\title{Competitive Distillation: A Simple Learning Strategy for \\ Improving Visual Classification}

\author{
Daqian Shi$^1$, Xiaolei Diao$^2$, Xu Chen$^1$, and Cédric M. John$^1$ \thanks{Corresponding author}\\
$^1$Digital Environment Research Institute (DERI), Queen Mary University of London, London, UK \\
$^2$School of Electronic Engineering and Computer Science, Queen Mary University of London, London, UK \\
{\tt\small \{d.shi, x.diao, xu.chen, cedric.john\}@qmul.ac.uk} 
}

\begin{document}
\maketitle

\begin{abstract}

Deep Neural Networks (DNNs) have significantly advanced the field of computer vision. To improve DNN training process, knowledge distillation methods demonstrate their effectiveness in accelerating network training by introducing a fixed learning direction from the teacher network to student networks. In this context, several distillation-based optimization strategies are proposed, e.g., deep mutual learning and self-distillation, as an attempt to achieve generic training performance enhancement through the cooperative training of multiple networks. However, such strategies achieve limited improvements due to the poor understanding of the impact of learning directions among networks across different iterations. In this paper, we propose a novel competitive distillation strategy that allows each network in a group to potentially act as a teacher based on its performance, enhancing the overall learning performance. Competitive distillation organizes a group of networks to perform a shared task and engage in competition, where competitive optimization is proposed to improve the parameter updating process. We further introduce stochastic perturbation in competitive distillation, aiming to motivate networks to induce mutations to achieve better visual representations and global optimum. The experimental results show that competitive distillation achieves promising performance in diverse tasks and datasets.

\end{abstract}

\section{Introduction}
\label{sec:Introduction}

Recent advancements in computer vision have been greatly facilitated by deep neural networks (DNNs), which have proven significant success in tasks such as image classification \cite{krizhevsky2017imagenet}, object detection \cite{redmon2016you}, and semantic segmentation \cite{zhang2018context}. Knowledge distillation exploits the deep representation capabilities of pre-trained DNNs, aiming to achieve effective network training with a lower cost \cite{parisotto2015actor}. By setting up a teacher-student learning paradigm, knowledge distillation allows a student network to learn from a teacher network by imitating its outputs \cite{hinton2015distilling}. Such a paradigm is considered a manifestation of collective intelligence \cite{leimeister2010collective}, which is a dynamic optimization strategy aimed at leveraging collaborative and competitive processes among multiple entities to generate complex decisions. The effectiveness of knowledge distillation methods demonstrates that learning by network collaboration is often an easier optimization task than directly learning from the target function \cite{ba2014deep}. 

\begin{figure}[!t]
	\centering
	\includegraphics[width=1\linewidth]{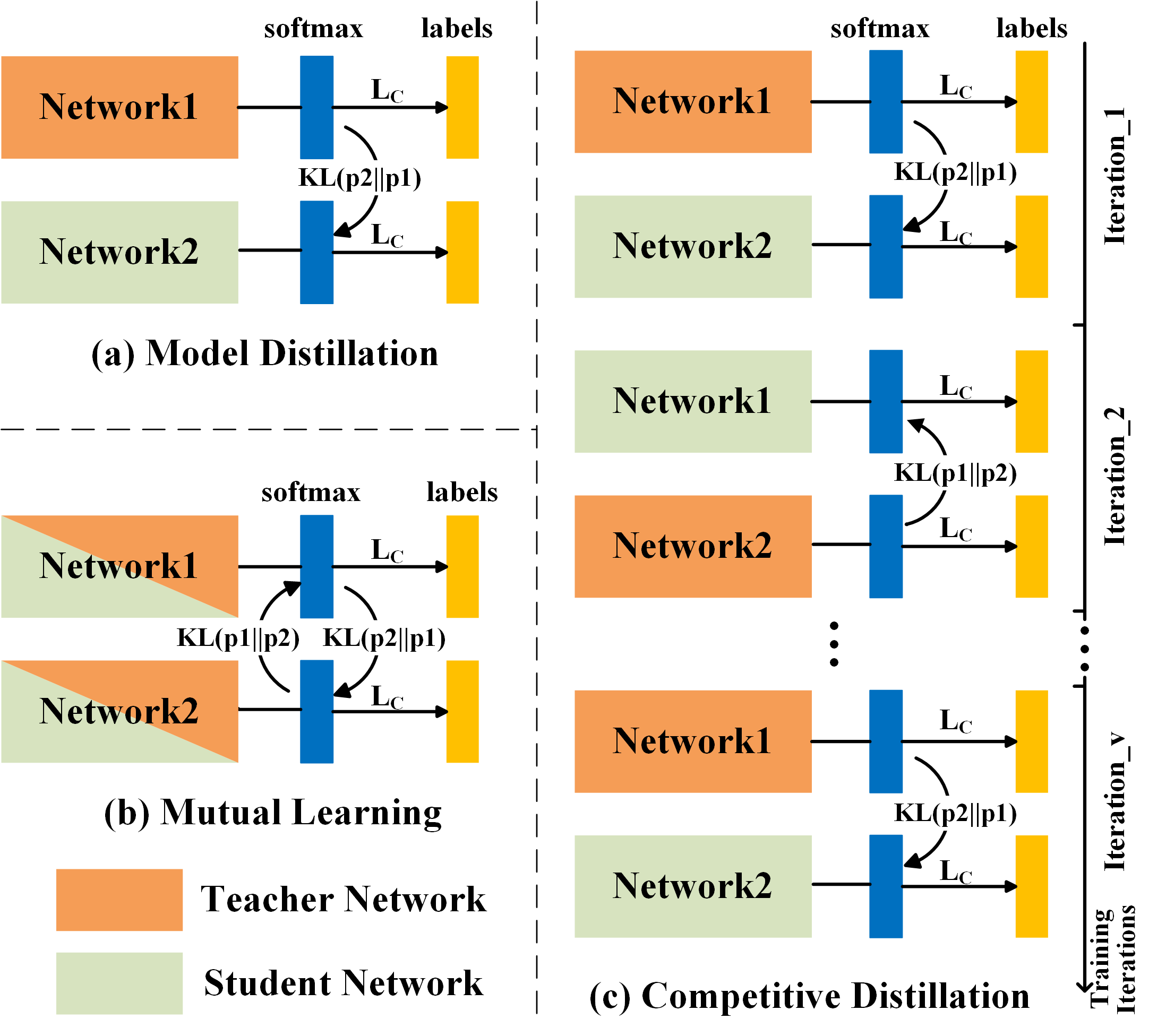}
    \caption{Comparison of parameter update strategies between general model distillation, mutual learning-based distillation, and proposed competitive distillation.
	\label{fig:1}}
\end{figure}


The most common cases of the teacher-student paradigm allow a student network to learn from a fixed teacher network, named model distillation, as shown in Fig.~\ref{fig:1}(a). The success of such methods in enhancing the efficiency of training student networks validates the rationality and effectiveness of the distillation process. Meanwhile, some distillation-based methods are specifically proposed to improve the overall task performance, e.g., deep mutual learning \cite{zhang2018deep} and self-distillation \cite{zhang2019your}. Fig.~\ref{fig:1}(b) demonstrates the structure of mutual learning-based methods, where each of the networks serves as both a teacher and a student network, enabling them to learn from each other by aligning each network's class posterior with the class probabilities of other networks during the learning iterations. Compared with training networks independently, mutual learning-based methods indeed enhance the overall task performance, aligning with the collaborative processes emphasized by collective intelligence strategies. However, it is reasonable for networks with poor performance to learn from superior ones, but counter-intuitive for the best-performing networks to learn from others. Thus, the learning direction of each network in a given iteration should be dynamically planned.

In this paper, we introduce a simple yet effective distillation-based strategy for improving visual classification, namely competitive distillation. Inspired by the competitive process from collective intelligence, we propose a competitive optimization-based parameter updating method to distill knowledge from the best-performing network to other networks at each training iteration dynamically. Specifically, the method involves a group of networks trained without a predetermined learning direction, whereby the winning network becomes the teacher for the current iteration and the others learn from the teacher as students, as shown in Fig.~\ref{fig:1}(c). Besides, inspired by genetic algorithm \cite{katoch2021review}, we propose a stochastic perturbation process to motivate networks to enhance competition for better visual representations. We randomly introduce image perturbations into one of the networks at each iteration, the aim is to induce mutations\footnote{Mutation is the operator to maintain the genetic diversity from one population to the next in the genetic algorithm.} \cite{mirjalili2019genetic} for providing the possibility for the network to escape local optima. The experimental results show that the proposed competitive distillation method achieves better performance than every single network when it learns individually in the traditional supervised learning scenario. In particular, the competitive distillation method outperforms state-of-the-art distillation methods on various visual tasks, including vision classification, person re-identification, and object detection. 
These results demonstrate that the competition optimization-based training strategy and stochastic perturbation process effectively improve the performance and efficiency of deep neural networks for various applications.
The contributions of our paper are summarized as follows:

\begin{itemize}
    \item We propose a novel training strategy, i.e., competitive distillation, for improving the performance of DNNs on various computer vision tasks, which exploits a competitive optimization strategy for effective parameter updating and knowledge distillation. 
    
    \item We introduce perturbations to our method by randomly introducing perturbed images on each network. Benefiting from the competitive optimization strategy, the positive perturbation results will be retained to provide the network with the possibility of breaking through local optima.

    
    \item We validate the proposed method on various image classification benchmarks. Compared to state-of-the-art distillation methods, our method achieves promising results in both task performance and training efficiency.
\end{itemize}

\section{Related Work}
\label{sec:Related Work}
\subsection{Knowledge Distillation}
Knowledge distillation has been proposed to open up new avenues for improving the efficiency and performance of deep neural networks \cite{hinton2015distilling}. 
Generally, these distillation-based methods \cite{romero2014fitnets, park2019relational, gou2021knowledge, wang2021knowledge} are conducted with a teacher-student framework that distills powerful large teacher networks into small student networks to reduce training costs. 
\cite{asif2019ensemble} introduces an ensemble knowledge distillation architecture that exploits multiple teacher networks to promote heterogeneity in feature learning. \cite{chen2022knowledge} attempt to narrow down the teacher-student performance gap by reusing a discriminative classifier from a pre-trained teacher network for student inference. 

To further improve the performance of trained networks, \cite{furlanello2018born} first exploited knowledge distillation as a network development method instead of a compression method, where they train student networks parameterized identically to their teachers.
This observation gives a hint that knowledge distillation is able to improve the performance of target networks, where researchers then introduce a strategy that implements the student network also as a teacher in the same distillation process, which means, one specific network can learn from others and also be distilled by others. 
For instance, deep mutual learning (DML) allows an ensemble of students to teach each other in a mutual distillation framework, which makes each student network improve compared with training them separately \cite{zhang2018deep}. 
Different from using multiple networks, \cite{zhang2019your} proposed a self-distillation method that makes only one network as both the teacher and student during the training process, allowing it to distill and learn from itself. Besides, Curriculum Temperature for Knowledge Distillation \cite{li2023curriculum} is proposed to control the task difficulty level during the training of students, aiming to improve performance by following an easy-to-hard curriculum.
\cite{gou2023multi} ask students to learn through multi-stages and self-reflect after each stage, aiming to enhance the ability of students to digest knowledge.
However, these methods do not consider the parameter updating priority during knowledge transformation, which affects the final performance of better-performing students.

\subsection{Collaborative Learning}

There are also other studies using multiple networks to collaborate for a single task. Generative adversarial networks (GANs) \cite{creswell2018generative, goodfellow2020generative}  are widely known as a framework consisting of discriminative networks and generative networks, for instance, \cite{28, shi2022charformer} introduced a two-step training method, where a discriminative noise estimator is trained to help with generating image pairs for training the generative denoiser. Recently, multi-modality networks \cite{zhou2019review} are also proposed to learn multiple networks jointly for the same task but in different domains, e.g., \cite{zhang2020emotion} proposed a multi-modal-based network for emotion recognition, which exploits both general photos and electroencephalograms by two separate networks.
Nested Collaborative Learning \cite{li2022nested} is proposed to learn representations more thoroughly by collaboratively learning multiple experts with different views together. By assigning different networks for training the same targets, collaborative learning for online knowledge distillation \cite{guo2020online, yang2023online} is proposed to emphasize the synergy among models, which applies online ensembling and network collaboration into a unified framework. However, ensembling does not always guarantee better results, performance fluctuations in some iterations can hinder network learning. Our proposed competitive distillation aims to enable the network to learn stably from superior performance through competition, thereby enhancing the overall performance of the target network.


\section{The Proposed Competitive Distillation}
\label{sec:Method}
\begin{figure*}[!t]
	\centering
	\includegraphics[width=1\linewidth]{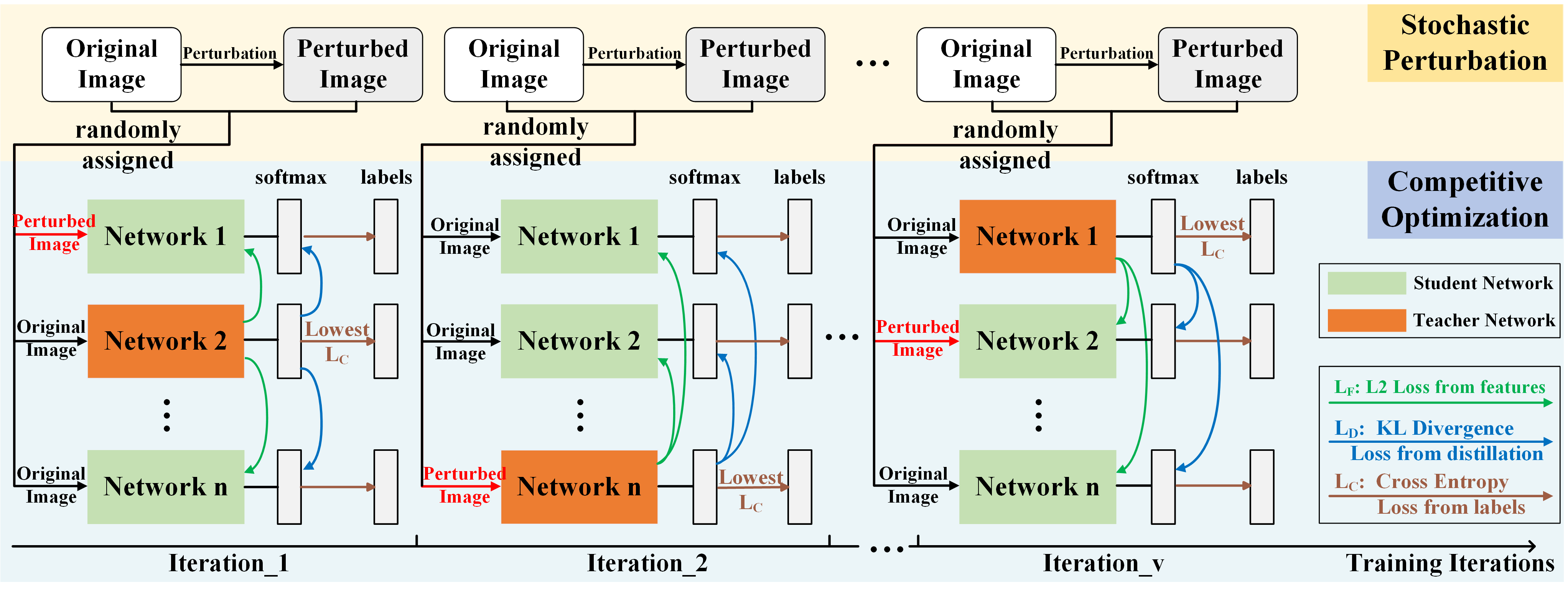}
    \caption{The overall framework of competitive distillation. We highlight the current best-performing network in orange, which will serve as the teacher network in the current iteration.
	\label{fig:2}}
\end{figure*}







Distillation-based network enhancing methods, e.g., DML, transfer knowledge between networks to improve overall task performance. Such methods can be considered a collaborative training procedure involving multiple networks \cite{zhang2018deep}. In this situation, assume we have a group of neural networks $\Theta$ that are trained for the same vision task, where $\Theta = \{\Theta_i | i = 1, 2, ..., n  \}, n \geq 2$, as at least two networks are required in these distillation-based methods. The network training can be modeled as follows: 
%
\begin{equation}
      \Theta^{t+1}_1, \Theta^{t+1}_2, ...,  \Theta^{t+1}_n = collaborate(\Theta^{t}_1, \Theta^{t}_2, ..., \Theta^{t}_n),
\label{equ:3.1}
\end{equation}
where we have $n$ networks $\Theta^{t}_1$, $\Theta^{t}_2$, and $\Theta^{t}_n$ 
trained in a collaborative training procedure $collaborate(\cdot)$, and $t$ refers to the training iteration of the network $\Theta_i$. Notice $collaborate(\cdot)$ will be specified as a detailed training algorithm in practice. As an improvement, our proposed strategy is intuitively derived from the dynamic optimization strategies of collective intelligence, which, in addition to utilizing network collaboration, also introduces a competitive process to lead networks toward better learning objectives. We define the training as follows: 
%
\begin{equation}
\begin{aligned}
 &     \Theta^{t+1}_1, \Theta^{t+1}_2, ...,  \Theta^{t+1}_n  \\ 
 &     =  collaborate(compete(\sigma (\Theta^{t}_1), \sigma (\Theta^{t}_2), ..., \sigma (\Theta^{t}_n))),
\end{aligned}
\label{equ:3.2}
\end{equation}
where $compete(\cdot)$ defines the competitive process, which identifies the optimal network to serve as the learning target for that iteration, $\sigma$ refers to the mutation variable that may be imposed on the network, intended to enhance the diversity of competition. Additionally, the visual classification task generally involves a training set $X$ with $M$ samples from $K$ classes, $ X = \left\{x_j \right\}^M _{j=1}$, the corresponding label set $Y$ is represented as $ Y = \left\{y_j \right\}^M _{j=1}$, where $y_j = 1, 2, ..., K$.
The probability that the neural network $\Theta_i$ recognizes that the sample $x_j$ belongs to the class $k$ can be expressed as:
\begin{equation}
      p^k_i(x_j) = \frac{exp(z^k_j)}{ \sum^N_{k=1} exp(z^k_j)},
\label{equ:1}
\end{equation}
where $z^k$ is the logit in network $\Theta_i$.  

Following the guide of Eq.~\ref{equ:3.2}, in our method, we propose competitive optimization that aims to enhance the network updating by setting the best-performing network as the teacher to guide other networks. The stochastic perturbation is designed to model the mutation variable $\sigma$ for improving visual representations and alleviating local optima issues. The overall framework for the proposed competitive distillation is shown in Fig.~\ref{fig:2}. Details are introduced in the following sections, and we summarize the complete processing steps of competitive distillation with Algorithm 1 in the supplementary documents.

\subsection{Competitive Optimization}
\label{sec:Optimization}

In competitive distillation, all networks are jointly optimized, and their optimization processes are closely interconnected. The key idea behind competitive distillation is to enable networks to compete with each other while engaging in collaborative training. This is specifically manifested by dynamically selecting the learning direction in each training iteration, namely distilling from the best-performing network to the other networks. 
The best-performing network emerges as the winner and serves as the teacher for the current iteration, while other networks act as students imitating the teacher, which we call competitive optimization. Specifically, in each iteration, the best-performing network is determined based on the cross-entropy loss $L_C$ by supervised classification predictions. For the ${t}^{th}$ iteration, we express the $i^{th}$ network as:
\begin{equation}
    \Theta^{t+1}_i = \left\{
\begin{tabular}{ll}
$\Theta^t_T$,  & if $L_C(\Theta^t_i) = Min(L_C(\Theta^t))$\\
$\Theta^t_S$,  & otherwise 
\end{tabular}\right.
\label{equ:3.1.1}
\end{equation}
where $\Theta^t_T$ and $\Theta^t_S$ represent the teacher and student network in the $t^{th}$ iteration, respectively. The network with the lowest loss is identified as the teacher, as the networks marked in orange in Fig.~\ref{fig:2}. Besides classification loss $L_C$, the student networks will be additionally updated based on the predictions of the teacher, where distillation loss $L_D$ and feature loss $L_F$ are applied. After determining the teacher and student networks for the ${(t+1)}^{th}$ iteration, we compute the stochastic gradients for all networks in group $\Theta$ and update network $\Theta_i$, as represented by:
\begin{equation}
    \Theta^{t+1}_i \gets \left\{
\begin{tabular}{ll}
 $\Theta^t_i - \gamma \frac{\partial L_{C^t_i}}{\partial \Theta^t_i}$,  & if $\Theta^t_i = \Theta^t_T$ \\
$\Theta^t_i - \gamma (\frac{\partial L_{C^t_i}}{\partial \Theta^t_i} +  \frac{\partial L_{D^t_i}}{\partial \Theta^t_i} + \frac{\partial L_{F^t_i}}{\partial \Theta^t_i})$,  & otherwise
\end{tabular}\right.
\label{equ:3.1.2}
\end{equation}
%
where $\gamma$ is the learning rate, and loss functions will be introduced in detail in Section 3.3. The competitive optimization-based parameter updating method is embedded in each batch-based parameter update step of the networks and runs throughout the entire training process. The optimization of all networks is iterative until convergence.


\begin{table*}[t!]
\centering
	\resizebox{1\linewidth}{!}{%
\begin{tabular}{c|cccccccccc}
\hline
Networks & ResNet-32 & ResNet-44 & ResNet-56 & ResNet-152 & MobileNet & WRN28-12 & WRN44-8 & Vim-S & ViT-B & Ceit-B \\ \hline
Parameters & 0.47M      & 0.66M      & 0.86M      & 60.2M       & 3.5M      & 36.5M           & 25.6M    & 26M      & 85.7M  & 88.6M   \\ \hline
\end{tabular}}
\caption{The number of parameters for involved networks.}
\label{tab:app2}
\end{table*}

\subsection{Stochastic Perturbation}

During the training process, the local optima issue is a common challenge 
\cite{bottou2012stochastic}. Inspired by genetic algorithm\footnote{Genetic algorithms are a type of classic optimization and search algorithm in collective intelligence.} \cite{lambora2019genetic}, we integrate the notion of mutation $\sigma$. Specifically, the implementation of alterations to the learning objects to obtain the comprehensive exploration of the space distribution and assist in escaping local optima. We propose a stochastic perturbation process to motivate the network training, wherein perturbations are applied to the input images of a randomly selected network at each iteration. Such perturbations aim to induce mutations and provide more significant gradient movements during training, thereby augmenting the likelihood of bypassing local optima, which benefits the network performance on broader data distribution in practice. Note that the mutation is a relatively aggressive method for distribution estimation that does not always positively affect training results in each iteration. Consequently, the stochastic perturbation process is devised in conjunction with the proposed competitive optimization strategy, with the objective of improving visual representations while alleviating the negative effects effectively. The intuition is that only the best-performing network is designated as the teacher, which allows for keeping the effective mutation since the mutation that effectively improves the teacher network performance will be transferred to the students during distillation. Meanwhile, the proposed strategy also prevents network training from negative mutations since the negatively affected networks are generally not the best-performing ones, namely students. As a result, the positive mutation will be preserved while the negative mutation will be ignored since it will not be learned by any networks.

We detail the stochastic perturbation processing as follows. Firstly, we set a pre-defined processing pool $P = \{P_r | r = 1,2,...,R\}$ that stores several physics-guided image processing functions as an approximate implementation of perturbation, including image fusion, splicing, noise injection, data deformation, and random cropping, etc, where $r$ represents the $r_{th}$ processing in the pool that contains $R$ functions. Notice that compared to normal data augmentation, we incorporated more aggressive processing parameters to introduce greater alterations to the original images, e.g., the scale of cropped region is set as [0.3, 0.7]. These perturbations in $P$ will be randomly selected to help update network parameters by considering different perturbation types and scales, which is similar to the motivation of mutation in genetic algorithms. Based on the perturbation pool $P$, we define the perturbation function $StochasticPerturb(\cdot)$ as a sequence of processing. The perturbation processing of the original image can be represented as follows:
\begin{equation}
    x^j_{Pert}, y^j_{Pert}  = StochasticPerturb(x^j_{Ori}, y^j_{Ori}),
\label{equ:3.2.1}
\end{equation}
where $x^j_{Ori}$  and $y^j_{Ori}$ are the original image and label, respectively. In this process, we stochasticlly select a perturbation $P_r$ from $P$ for transforming $x^j_{Ori}$ into the perturbed image $x^j_{Pert}$, and obtain the label $y^j_{Pert}$ correspondingly.

\subsection{Loss Functions}
\label{sec:loss}
To determine the best-performing network at each iteration and select the teacher networks, we compare them based on the supervised learning loss $L_C$, which is the cross-entropy error between the predicted and ground-truth labels, as: 
\begin{equation}
      L_{C_i} = - \sum^M_{j=1} \sum^K_{k=1} y_{jk} log(p^k_j(x_j)).
\label{equ:2}
\end{equation}
The $i^{th}$ network with the minimum $L_{C_i}$ is considered the best-performing network and serves as the teacher in the current iteration, denoted by the orange color in Fig.~\ref{fig:2}, while the other networks will act as students, denoted by the green color. To enhance the performance of the student networks, the teacher network provides training experience to the students through two forms of constraints: distillation loss $L_D$ and feature loss $L_F$. 

Specifically, we introduce Kullback-Leibler (KL) divergence \cite{joyce2011kullback}  as $L_D$ to quantify the match between the predictions $p_i$ and $p_t$ (represented student and teacher, respectively) of every two networks, with the goal of making the student approximate the teacher. It is computed between the outputs of the softmax layer. The KL distance of $i^{th}$ network from $p_i$ to $p_t$ can be expressed as:
\begin{equation}
      L_{D_i} =  D_{KL}(p_t \Vert p_i) =  \sum^M_{j=1} \sum^K_{k=1} p^m_t (x_j) log \frac{ p^m_t (x_j)}{ p^m_i (x_j) }.
\label{equ:3}
\end{equation}
Furthermore, we introduce $L_F$ to guide the student network to learn the features extracted by the teacher network \cite{romero2014fitnets}. It is obtained by computing the L2 loss between the feature maps of the student and the teacher network. Through the L2 loss, the implicit knowledge in the feature maps extracted by the teacher is transferred to each student network, which can be expressed as:
\begin{equation}
     L_{F_i} = \sum^M_{j=1} \Vert F_i – F_t \Vert^2_2,
\label{equ:4}
\end{equation}
where $F_i$ and $F_t$ represent the features in the student network $\theta_i$ and the features in the teacher network $\theta_t$, respectively. Thus, the overall loss function for each student network can be expressed as:
\begin{equation}
     L^S_{\theta_i } = L_{C_i} + \alpha L_{D_i} + \beta L_{F_i},
\label{equ:5}
\end{equation}
where $\alpha$ and $\beta$ are two hyperparameters used to balance different losses. While for the teacher network,  $\alpha$ and $\beta$ are zero, which means that the supervision of the teacher network just comes from labels. Thus, its loss is composed of classification loss, as:
\begin{equation}
     L^T_{\theta_i} = L_{C_i}.
\label{equ:6}
\end{equation}
It is worth noting that in competitive distillation, the roles of ``teacher'' and ``student'' networks are dynamically designated in each iteration. Every network is competing to become the teacher, and knowledge will be transferred to other student networks, while the teacher network itself will also be updated in each iteration.

\section{Experiments and Discussions}
\label{sec:Experiments}

\subsection{Datasets and Settings}

\begin{table*}[t]
\centering
	\resizebox{1\linewidth}{!}{%
\begin{tabular}{@{}cc|ccccc|ccccc|ccccc@{}}
\toprule
\multicolumn{2}{c|}{\multirow{3}{*}{Networks}} & \multicolumn{5}{c|}{CIFAR10}                                                                         & \multicolumn{5}{c|}{CIFAR100}                                                                        & \multicolumn{5}{c}{ImageNet}                                                                                                                                                       \\ \cmidrule(l){3-17} 
\multicolumn{2}{c|}{}                          & \multirow{2}{*}{Ind.} & \multirow{2}{*}{DML} & \multicolumn{3}{c|}{Competitive Distillation}  & \multirow{2}{*}{Ind.} & \multirow{2}{*}{DML} & \multicolumn{3}{c|}{Competitive Distillation}  & \multicolumn{1}{c}{\multirow{2}{*}{Ind.}} & \multicolumn{1}{c}{\multirow{2}{*}{DML}} & \multicolumn{3}{c}{Competitive Distillation}                                         \\ \cmidrule(lr){5-7} \cmidrule(lr){10-12} \cmidrule(l){15-17} 
\multicolumn{2}{c|}{}                          &                              &                      & Acc.  & Imp-Ind & \multicolumn{1}{c|}{Imp-DML} &                              &                      & Acc.  & Imp-Ind & \multicolumn{1}{c|}{Imp-DML} & \multicolumn{1}{c}{}                             & \multicolumn{1}{c}{}                     & \multicolumn{1}{c}{Acc.} & \multicolumn{1}{c}{Imp-Ind} & \multicolumn{1}{c}{Imp-DML} \\ \midrule \hline
Net 1                 & \multicolumn{1}{c|}{ResNet-32}                 & 91.98 & 92.36 & 92.95 & 0.97 & 0.59 & 68.73 & 70.87 & \textbf{71.45} & 2.72 & 0.58 & 72.86 & 73.65 & \textbf{74.35} & 1.49 & 0.70 \\
Net 2                 & \multicolumn{1}{c|}{ResNet-32}                 & 91.98 & 92.67 & \textbf{93.05} & 1.07 & 0.38 & 68.73 & 70.83 & 71.36 & 2.63 & 0.53 & 72.86 & 74.07 & 74.27 & 1.41 & 0.20 \\ \midrule

Net 1                 & \multicolumn{1}{c|}{MobileNet}                 & 93.63 & 94.15 & 94.45 & 0.82 & 0.30 & 73.59 & 75.67 & \textbf{76.22} & 2.63 & 0.55 & 74.34 & 74.96 & 75.51 & 1.17 & 0.55 \\
        Net 2                 & \multicolumn{1}{c|}{MobileNet}                 & 93.63 & 94.28 & \textbf{94.52} & 0.89 & 0.24 & 73.59 & 76.13 & 76.19 & 2.60 & 0.06 & 74.34 & 75.28 & \textbf{75.55} & 1.21 & 0.27 \\ \midrule


Net 1                 & \multicolumn{1}{c|}{ResNet-44}                 & 93.30 & 93.66 & 94.02 & 0.72 & 0.36 & 72.26 & 74.96 & 75.21 & 2.95 & 0.25 & 75.96 & 77.98 & \textbf{78.08} & 2.12 & 0.10 \\
Net 2                 & \multicolumn{1}{c|}{MobileNet}                 & 93.63 & 94.32 & \textbf{94.53} & 0.90 & 0.21 & 73.59 & 76.05 & \textbf{76.61} & 3.02 & 0.52 & 74.34 & 75.66 & 75.59 & 1.25 & -0.07 \\ \midrule


Net 1              & \multicolumn{1}{c|}{ResNet-56}                    & 94.02 & 94.69 & 94.65 & 0.63 & -0.04 & 73.76 & 75.31 & 76.76 & 3.00 & 1.45 & 77.05 & 78.48 & 78.97 & 1.92 & 0.49 \\
Net 2              & \multicolumn{1}{c|}{ResNet-152}                   & 94.43 & 94.81 & \textbf{94.97} & 0.54 & 0.16 & 76.66 & 78.64 & \textbf{79.80} & 3.14 & 1.16 & 77.23 & 79.31 & \textbf{79.85} & 2.62 & 0.54 \\ \midrule

Net 1                 & \multicolumn{1}{c|}{CeiT-B}                      & 96.60 & 97.03 & \textbf{97.44} & 0.84 & 0.41 & 86.28 & 87.77 & \textbf{87.79} & 1.51 & 0.02 & 85.47 & 85.79 & 86.08 & 0.61 & 0.29 \\
Net 2                 & \multicolumn{1}{c|}{ViT-B}                                            & 96.19 & 96.80 & 97.05 & 0.86 & 0.25 & 85.73 & 86.48 & 86.98 & 1.25 & 0.50 & 84.25 & 85.34 & \textbf{86.27} & 2.02 & 0.93 \\ \midrule

Net 1                 & \multicolumn{1}{c|}{CeiT-B}                      & 96.60 & 97.22 & 97.28 & 0.68 & 0.06 & 86.28 & 87.58 & \textbf{88.30} & 2.02 & 0.72 & 85.47 & 86.11 & \textbf{86.56} & 1.09 & 0.45 \\
Net 2                 & \multicolumn{1}{c|}{CeiT-B}                      & 96.60 & 96.93 & \textbf{97.40} & 0.80 & 0.47 & 86.28 & 87.76 & 88.05 & 1.77 & 0.29 & 85.47 & 86.17 & 86.33 & 0.86 & 0.16 \\  \midrule 
Net 1                 & \multicolumn{1}{c|}{Vim-S}      & 94.68 & 95.32 & 95.53          & 0.85 & 0.21 & 84.52 & 85.61 & 86.24          & 1.72 & 0.63 & 80.46 & 81.23 & 81.83          & 1.37 & 0.60                 \\
Net 2                 & \multicolumn{1}{c|}{Vim-S}   & 94.68 & 95.58 & \textbf{95.96} & 1.28 & 0.38 & 84.52 & 85.86 & \textbf{86.43} & 1.91 & 0.57 & 80.46 & 81.37 & \textbf{82.04} & 1.58 & 0.67          \\  \midrule \hline
Net 1              & \multicolumn{1}{c|}{ResNet-32}                    & 91.98 & 92.94 & 93.18 & 1.20 & 0.24 & 68.73 & 71.07 & 71.45 & 2.72 & 0.38 & 72.86 & 73.48 & 74.42 & 1.56 & 0.94 \\
Net 2              & \multicolumn{1}{c|}{WRN28-12}              & 96.45 & 96.88 & \textbf{97.21} & 0.76 & 0.33 & 77.57 & 78.62 & \textbf{79.33} & 1.76 & 0.71 & 78.46 & 79.56 & \textbf{80.24} & 1.78 & 0.68 \\
Net 3              & \multicolumn{1}{c|}{MobileNet}                    & 93.63 & 94.58 & 94.92 & 1.29 & 0.34 & 73.59 & 75.82 & 76.15 & 2.56 & 0.33 & 74.34 & 75.19 & 75.76 & 1.42 & 0.57 \\ \midrule

Net 1              & \multicolumn{1}{c|}{ResNet-32}                    & 91.98 & 92.55 & 92.94 & 0.96 & 0.39 & 68.73 & 70.26 & 70.44 & 1.71 & 0.18 & 72.86 & 73.21 & 74.26 & 1.40 & 1.05 \\
Net 2              & \multicolumn{1}{c|}{ResNet-32}                    & 91.98 & 92.70 & 93.16 & 1.18 & 0.46 & 68.73 & 69.95 & 71.21 & 2.48 & 1.26 & 72.86 & 72.90 & \textbf{74.59} & 1.73 & 1.69 \\
Net 3              & \multicolumn{1}{c|}{ResNet-32}                    & 91.98 & 92.78 & \textbf{93.32} & 1.34 & 0.54 & 68.73 & 70.84 & \textbf{71.49} & 2.76 & 0.65 & 72.86 & 73.19 & 74.08 & 1.22 & 0.89 \\ \midrule

Net 1              & \multicolumn{1}{c|}{ResNet-32}                    & 91.98 & 93.08 & 93.08 & 1.10 & 0.00 & 68.73 & 69.75 & 71.56 & 2.83 & 1.81 & 72.86 & 74.37 & 74.39 & 1.53 & 0.02 \\
Net 2              & \multicolumn{1}{c|}{ResNet-44}                    & 93.30 & 93.54 & 93.73 & 0.43 & 0.19 & 72.26 & 73.99 & 75.48 & 3.22 & 1.49 & 75.96 & 76.56 & 78.10 & 2.14 & 1.54 \\
Net 3              & \multicolumn{1}{c|}{ResNet-56}                    & 94.02 & 94.42 & \textbf{94.87} & 0.85 & 0.45 & 73.76 & 75.06 & \textbf{76.69} & 2.93 & 1.63 & 77.05 & 77.47 & \textbf{78.94} & 1.89 & 1.47 \\
\bottomrule               
\end{tabular}}
\caption{Comparison of the Top-1 accuracy (\%) of different network groups (when $n=2$ or $n=3$) on the three dataset.  ``Ind.'' represents the accuracy of networks based on independent training. ``DML'' represents the accuracy of networks obtained by deep mutual learning\cite{zhang2018deep}. ``Acc.'' denotes the accuracy obtained by competitive distillation. ``Imp-Ind'' measures the difference in accuracy between the network learned with competitive distillation and the same network learned independently. ``Imp-DML'' measures the difference in accuracy between the network learned with competitive distillation and deep mutual learning.}
\label{tab:1}
\end{table*}

\textbf{Datasets.} 
We evaluate the proposed competitive distillation strategy on four datasets CIFAR10 \cite{krizhevsky2010cifar}, CIFAR100 \cite{krizhevsky2010cifar}, ImageNet \cite{russakovsky2015imagenet} and Market-1501 \cite{zheng2015scalable}. The details are as follows\footnote{Note that we also evaluate the competitive optimization on more tasks for a comprehensive validation, for more details please find the sup doc.}:

\begin{itemize}

    \item \textbf{CIFAR10} \& \textbf{CIFAR100}: Both datasets consist of 32×32 pixel RGB images, with CIFAR10 containing 10 classes and CIFAR100 comprising 100 classes. Each dataset includes 50K training images and 10K test images.

    \item \textbf{ImageNet}: We choose the version ILSVRC2012 \cite{russakovsky2015imagenet}, which includes 1000 classes. Each class contains 1000+ images, with a total of approximately 1.2 million images used for training and 50k images for validation. We resize them into 256$\times$256 pixels RGB images.

    \item \textbf{Market-1501}: A large dataset for person re-identification (Re-ID), containing 32,668 images of 1,501 identities captured across six camera views. The dataset is divided into 751 identities for training and 750 for testing. 



\end{itemize}

\noindent \textbf{Networks.} 
%
To validate the effectiveness of the competitive distillation strategy across networks of different scales, we considered networks with different types and scales in our experiments. These include compact networks of typical student scales, e.g., ResNet-32, ResNet-44, ResNet-56 \cite{he2016deep}, and MobileNet \cite{howard2017mobilenets}; as well as large networks of typical teacher scales, including ResNet-152, Wide-ResNet (WRN) \cite{hounie2022automatic} such as WRN28-12, WRN44-8, vision mamba Vim-small (Vim-S) \cite{vim2024} and transformer-based networks ViT-B \cite{dosovitskiy2020image} and Ceit-B \cite{yuan2021incorporating}. To provide a clearer understanding of the network scales, we present the number of parameters for all the networks we employed in Table~\ref{tab:app2}.

\noindent \textbf{Implementation details.} 
We implement all the networks and training procedures in PyTorch, and we conduct all experiments with eight NVIDIA A100 GPUs. During the network training, we use learning rate decay, L2 regularization, and classic data augmentation techniques. Specifically, we utilize Stochastic Gradient Descent (SGD) with Nesterov momentum \cite{bottou2012stochastic}, an initial learning rate of 0.1, momentum set to 0.9, and batch size set to 128. 
The loss hyperparameters $\alpha$ and $\beta$ of the student network are set to 1 in the experiments.
The evaluation metric used for all datasets was reported as the Top-1 classification accuracy. 

\begin{table*}[t]
\centering
\resizebox{1\linewidth}{!}{%
\begin{tabular}{@{}c|cc|ccccccccccccc@{}}
\toprule
\multirow{2}{*}{Datasets} & \multicolumn{2}{c|}{\multirow{2}{*}{Network Types}} & \multicolumn{2}{c}{Baseline}      & KD                          & AT                         & PESF-    & SD    & NKD    & \multicolumn{2}{c}{DML}                  & \multicolumn{2}{c}{L-MCL}                 & \multicolumn{2}{c}{Competitive } \\
\multicolumn{1}{c|}{} & \multicolumn{2}{c|}{}                               & \multicolumn{2}{c}{(Independent)} & \cite{hinton2015distilling} & \cite{zagoruyko2016paying} & KD\cite{rao2023parameter}    & \cite{zhang2019your}  & \cite{yang2023knowledge}    & \multicolumn{2}{c}{\cite{zhang2018deep}} & \multicolumn{2}{c}{\cite{yang2023online}} & \multicolumn{2}{c}{Distillation (Ours)} \\ \midrule
\multirow{6}{*}{CIFAR100} & \textcolor{myred}{Net1}                     & \textcolor{mygreen}{Net2}                     & \textcolor{myred}{Net1}             & \textcolor{mygreen}{Net2}             & \textcolor{myred}{Net1}           & \textcolor{myred}{Net1}             & \textcolor{myred}{Net1}             & \textcolor{myred}{Net1}                       & \textcolor{myred}{Net1}             & \textcolor{myred}{Net1}                & \textcolor{mygreen}{Net2}                & \textcolor{myred}{Net1}                & \textcolor{mygreen}{Net2}                 & \textcolor{myred}{Net1}                  & \textcolor{mygreen}{Net2}                   \\
& ResNet-32                  & MobileNet                & 68.73 & 73.59 & 68.89 & 69.75 & 70.72 & 70.65 & 70.35 & 70.94 & 76.09 & 70.78 & 76.17 & \textbf{71.42} & \textbf{76.34} \\
& ResNet-32                & ResNet-56                & 68.73 & 73.76 & 68.57 & 69.88 & 71.08 & 71.59 & 70.94 & 71.19 & 75.88 & 71.11 & 75.50 & \textbf{71.37} & \textbf{76.23}\\
& ResNet-56               & ResNet-152                & 73.76 & 76.66 & 74.51 & 74.91 & 74.93 & 75.79 & 75.12 & 75.31 & 78.64 & 75.03 & 78.84 & \textbf{76.76} & \textbf{79.80} \\

& ResNet-56          & WRN28-12                & 73.76 & 77.57 & 74.60 & 74.33 & 75.85 & 76.07 & 75.59 & 75.96 & 79.12 & 75.18 & 78.76 & \textbf{76.63} & \textbf{79.29} \\
& ResNet-152           & WRN44-8               & 76.66 & 80.12 & 77.18 & 76.90 & 77.32 & 78.14 & 77.10 & 79.08 & 81.25 & 77.64 & 80.61 & \textbf{79.47} & \textbf{81.67} \\
& WRN28-12           & WRN44-8          & 77.57 & 80.12 & 77.95 & 78.54 & 78.80 & 78.84 & 78.62 & 79.39 & 81.41 & 79.41 & 80.36 & \textbf{79.82} & \textbf{81.82} \\

\midrule
\multirow{3}{*}{Market-1501} & ResNet-32 & MobileNet    & 81.07 & 85.01 & 82.14 & 82.45 & 82.89 & 83.56 & 83.12 & 84.43 & 89.34 & 84.02 & 88.54 & \textbf{84.65} & \textbf{89.82} \\
                             & ResNet-32 & ResNet-44  & 81.07 & 85.66 & 82.33 & 83.16 & 83.47 & 83.75 & 82.89 & 84.75 & 89.97 & 84.27 & 89.15 & \textbf{84.86} & \textbf{90.03} \\
                             & ResNet-56 & ResNet-152 & 88.57 & 94.77 & 89.01 & 89.34 & 89.76 & 89.48 & 90.03 & 90.36 & 95.64 & 90.43 & 95.32 & \textbf{91.28} & \textbf{96.17} \\ \bottomrule
\end{tabular}}
\caption{Comparison with distillation-based methods on CIFAR100 and Market-1501, using Top-1 accuracy (\%) and Rank-1 (\%), respectively. Net1 and Net2 are highlighted in \textcolor{myred}{red} and \textcolor{mygreen}{green}, respectively. The best results for each network are marked by \textbf{bold}.}
\label{tab:3}
\end{table*}

\begin{table*}[t]
\centering
\resizebox{1\linewidth}{!}{%
\begin{tabular}{@{}c|cccc|cccc@{}}
\toprule
\multirow{2}{*}{Net1/Net2}     & \multicolumn{4}{c|}{Competitive   Optimization}             & \multicolumn{4}{c}{DML \cite{zhang2018deep}}     \\ \cmidrule(r){2-5}  \cmidrule(l){6-9} 
        & Backbone & Backbone+$L_F$ & Backbone+Pert & Backbone+$L_F$+Pert & Backbone & Backbone+$L_F$ & Backbone+Pert & Backbone+$L_F$+Pert \\ \midrule
ResNet-32/ResNet-32                                 & 71.13    & 71.24         & 71.37         & 71.45              & 70.87    & 71.09         & 70.92         & 71.09              \\
MobileNet/ResNet-44                                 & 76.24    & 76.32         & 76.48         & 76.61              & 76.09    & 76.20         & 75.78         & 75.53              \\
ResNet-32/ResNet-152                                & 71.38    & 71.52         & 71.67         & 71.84              & 71.06    & 71.11         & 70.86         & 70.73              \\
ResNet-56/ResNet-152                                & 76.12    & 76.43         & 76.65         & 76.76              & 75.31    & 75.72         & 76.18         & 76.38              \\
CeiT-B/CeiT-B                                       & 88.03    & 88.12         & 88.28         & 88.30              & 87.58    & 87.84         & 87.65         & 87.70              \\ \bottomrule
\end{tabular}}
\caption{Ablation studies on the feature loss function $L_F$ and stochastic perturbation for competitive optimization strategy and DML on CIFAR100, the presented results are from Net1 in each group, using Top-1 accuracy  (\%).} 
\label{tab:5}
\end{table*}

\subsection{Experiment Results}
\label{subsec:experiment-results}

We conducted independent training and group learning of various networks on three datasets: CIFAR10, CIFAR100, and ImageNet. The learning strategies included DML-based training and competitive distillation. Table~\ref{tab:1} presents the results of group experiments performed by two networks, along with the accuracy improvements achieved by competitive distillation compared to training independently and DML. Based on the observations of Table~\ref{tab:1}, we can draw the following conclusions:
(i) Group learning outperforms individual learning, as indicated by the positive values in all ``Imp-Ind'' columns. This demonstrates that competitive distillation enhances network performance compared to independent learning.
(ii) Competitive distillation usually brings more improvements than mutual learning, where most of the values in the ``Imp-DML'' column are positive. This suggests that constraining the network to approach a better-performing network in competitive distillation can transfer knowledge effectively.
(iii) The proposed strategy brings improvements on both networks, which demonstrates the teacher network is designated dynamically and knowledge transformation happens on both networks, even under the training setting with a large difference in network scales. 
(iv) Considering the network performance when trained independently, a network trained with a better-performing network usually benefits more.
(v) Using different types and scale of networks, including CNN-based, Transformer-based, and Mamba-based networks, leads to performance improvements, suggesting that competitive distillation is a generic parameter updating strategy. Compared to independent training, our method achieves an average accuracy improvement of 0.91\%, 2.25\%, and 1.58\% on the three datasets, respectively.

In order to gain further insight into the effectiveness of competitive distillation, we conducted group learning experiments using more networks. The experimental results are presented in the last three sets of experiments in Table~\ref{tab:1}, and more results from experiments with multiple networks can be found in the supplementary document. We observed that competitive distillation presents a similar trend in multiple network groups, consistently outperforming independent training and mutual learning strategies on almost all benchmarks. Specifically, for a network, collaborative distillation exhibits an improvement in average accuracy by 1.69\% compared to independent training, and 0.59\% compared to deep mutual learning, regardless of the number of network configurations for collaboration. For example, when training two, three, and four ResNet-32 models collaboratively, the average accuracy of ResNet-32 increased by 1.72\%, 1.64\%, and 1.76\%, respectively, compared to independent training. Compared to DML, the accuracy gains were 0.50\%, 0.79\%, and 0.40\%, respectively.

\noindent \textbf{Comparison with distillation-based methods}. 
To compare different distillation methods, we conducted experiments on CIFAR100 and Market-1501, corresponding to visual classification and Re-ID tasks, respectively. The results are presented in Table \ref{tab:3}. We independently trained two networks, Net1 and Net2, and compared their performance with the corresponding networks trained under different distillation algorithms to observe performance improvements. For classical distillation methods, including Knowledge Distillation (KD) \cite{hinton2015distilling}, Attention Transfer (AT) \cite{komodakis2017paying}, and PESF-KD \cite{rao2023parameter}, we set Net1 as the student network and Net2 as the teacher network. Since these distillation methods update only the student network’s parameters, we report the performance of the student network, Net1, for each method. Additionally, self-distillation algorithms SD \cite{zhang2019your} and NKD \cite{yang2023knowledge} employ only a single network for distillation. Therefore, we used Net1 as the training backbone to report the self-distillation results. Group leaning methods like DML \cite{zhang2018deep}, online distillation L-MCL \cite{yang2023online}, and our proposed competitive distillation involve parameter updates for both Net1 and Net2. Consequently, we report the best-performing results among all trained networks for these methods. The results on two tasks exhibit similar trends. First, we observe that distillation algorithms enable the student network, Net1, to achieve performance comparable to or even better than that of independently trained models. Moreover, after applying self-distillation methods, DML and L-MCL, the final performance of Net2 significantly surpasses that of its independently trained counterpart. Most importantly, we observe that our proposed competitive distillation strategy significantly outperforms other methods on both networks. This finding suggests that learning from a better-performing network facilitates a more effective transfer of classification-supportive features, further validating the effectiveness of our method.

\subsection{Ablation Study}

\noindent \textbf{Impact of stochastic perturbation.}
To validate the effectiveness of perturbation, we introduce perturbations on backbones in both DML and competitive optimization strategies. In Table~\ref{tab:5}, we execute the group training and report the results of the Net1 of each group. We have two main observations. (i) Stochastic perturbation acts positively when applied on competitive distillation, where the performance of each group improves stably, including the Backbone+Pert and Backbone+$L_F$+Pert. (ii) Perturbations do not improve DML significantly, which indicates that perturbation, as an aggressive distribution estimation method, may limit network training. (iii) The backbone of competitive distillation outperforms DML, which means the proposed group training strategy is a generic improvement. Thus, this observation proves the effectiveness of the stochastic perturbation method used in conjunction with the competitive optimization strategy, which improves the performance of knowledge transformation. 

\noindent \textbf{Impact of the loss functions.}
We also assessed the effectiveness of feature loss $L_F$. The experimental settings are similar to the assessment of stochastic perturbation, and results are also shown in Table~\ref{tab:5}. We observe stable improvements to some extent in both competitive distillation and DML training groups by using feature loss, which shows the effectiveness of using feature loss $L_F$ in our proposed method. More importantly, our method consistently outperforms DML even under the same loss function and perturbation conditions, further demonstrating the effectiveness of the proposed competitive optimization strategy.


\subsection{Discussions}

\begin{figure}[!t]
	\centering
	\includegraphics[width=1\linewidth]{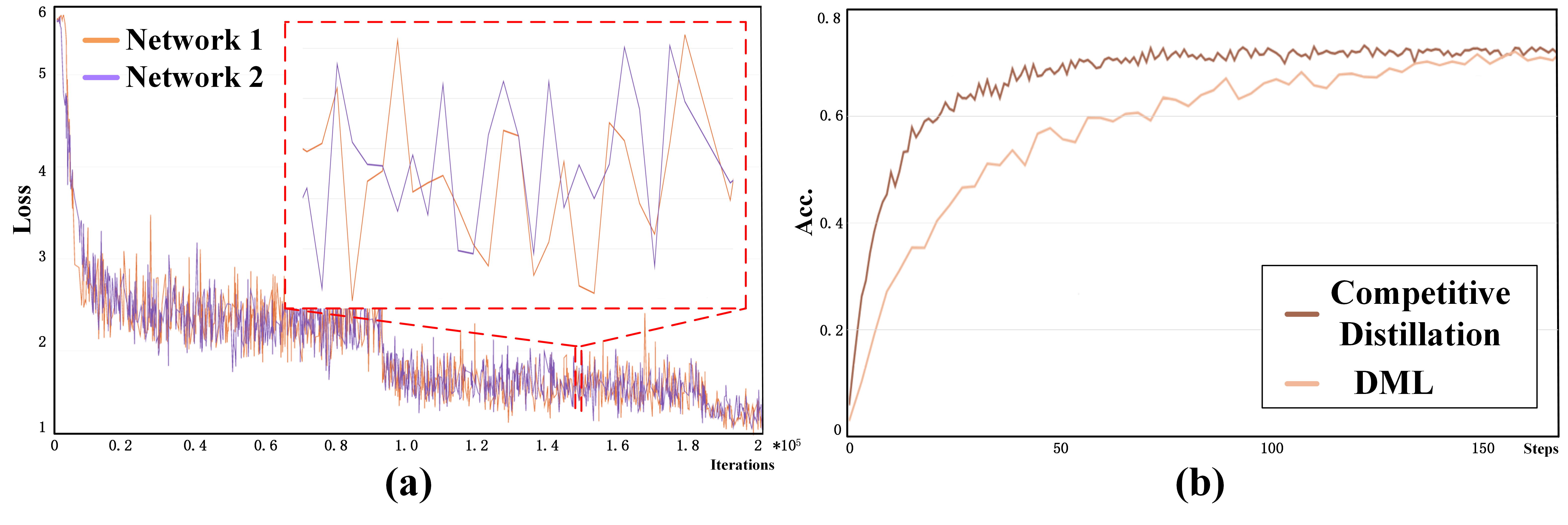}
    \caption{(a) The cross-entropy loss $L_{C_i}$ of each network in competitive distillation as it changes with iterations during the training process. (b) Display the accuracy and convergence of competitive distillation and DML during training.
	\label{fig:3}}
 \vspace{-0.4cm}
\end{figure}

\textbf{The demonstration of ``competitive''.}
To further explore the competition during network training in the competitive distillation method, we present a case study in Fig.~\ref{fig:3}(a) of training two networks to track the changes in the cross-entropy loss $L_{C_i}$, where Network 1 is a ResNet-56 and Network 2 is a ResNet-152. We observed that Network 1 and Network 2 alternately emerge as the network with lower $L_{C_i}$, indicating that the teacher network is continually switching, with Networks 1 and 2 alternating roles as teacher and student during training. Moreover, we also observe that the alternation of the teacher network emerges in networks with significant scale differences. These observations, along with results from Tables~\ref{tab:1}, demonstrate the dynamic designation of teacher networks and knowledge transfer between networks, aligning with our intuition behind the design of competitive distillation.

\noindent \textbf{Convergence efficiency.}
We observed that competitive distillation typically converges in fewer steps compared to DML, as illustrated in Fig.~\ref{fig:3}(b). On the CIFAR100 dataset, competitive distillation often converges within 100 steps, whereas DML typically requires 150 steps. We also measured and found that when training the same number of networks for competitive distillation and DML, the average time cost for each step is less or the same. This observation indicates the effectiveness of transferring knowledge of our proposed learning strategy and also indicates that proper knowledge transformation priority will contribute to reducing training time.

\noindent \textbf{Limitations.}
The proposed competitive distillation aims to enhance the overall performance of neural networks on vision tasks. Although it achieves higher efficiency and lower computational cost compared to multi-network training methods such as DML and L-MCL, it still requires more computational resources than training a single network individually. This makes the method more suitable for use in a pre-trained environment, with limited flexibility in real-time or low-latency applications.

\section{Conclusion}
\label{sec:Conclusion}

In this paper, we propose a novel network learning strategy to improve the overall performance of visual classification tasks, namely competitive distillation. Specifically, a competitive optimization method is proposed to engage a group of networks in competition to facilitate knowledge transfer, and the best-performing network guides network training in each iteration. Furthermore, we introduce stochastic perturbation for competitive distillation, aiming to motivate networks to induce mutations to achieve better visual representations and global optima. The experimental results show that competitive distillation achieves promising performance with higher convergence efficiency. 
As a generic parameter updating strategy, our future work will focus on adapting competitive distillation to NLP fields and solving the mentioned limitations. 



{
    \small
    \bibliographystyle{ieeenat_fullname}
    \bibliography{main}
}

\end{document}